\pdfoutput=1

\documentclass[11pt]{article}

\usepackage[]{nlp}

\usepackage{times}
\usepackage{latexsym}

\usepackage[T1]{fontenc}

\usepackage[utf8]{inputenc}

\usepackage{microtype}
\usepackage{booktabs}
\usepackage{xspace}
\usepackage{graphicx}
\usepackage{mkolar_definitions}
\usepackage{rotating}
\usepackage{multirow}
\usepackage{mathtools}

\newcommand{\entity}{\ensuremath{e}}
\newcommand{\mention}{\ensuremath{m}}
\newcommand{\mentions}{\ensuremath{\Mcal}}
\newcommand{\doc}{\ensuremath{d}}
\newcommand{\docs}{\ensuremath{\Dcal}}
\newcommand{\KB}{\ensuremath{\Ecal}}
\newcommand{\cluster}{\ensuremath{c}}
\newcommand{\clusters}{\ensuremath{\Ccal}}
\newcommand{\NIL}{\ensuremath{\textsc{nil}}}
\newcommand{\mentionmention}{\ensuremath{\phi}}
\newcommand{\mentionentity}{\ensuremath{\psi}}
\newcommand{\edges}{\ensuremath{E}}

\newcommand{\graph}{\ensuremath{G}}
\newcommand{\vertices}{\ensuremath{V}}
\newcommand{\zeshel}{ZeShEL\xspace}

\newcommand{\independent}{\textsc{Independent}\xspace}

\newcommand{\clusterbased}{\textsc{Clustering-based}\xspace}
\newcommand{\ours}{\textsc{Arborescence-based}\xspace}
\newcommand{\inBatch}{\textsc{In-Batch Negatives}\xspace}
\newcommand{\knnNegs}{\textsc{k-NN Negatives}\xspace}

\newcommand{\mentionencoder}{\ensuremath{\mathtt{Enc}_\mathrm{M}}}
\newcommand{\entityencoder}{\ensuremath{\mathtt{Enc}_\mathrm{E}}}

\newcommand{\minibatch}{\ensuremath{B}}

\newcommand{\hardnegatives}{k-NN\xspace}
\newcommand{\inbatch}{In-batch\xspace}

\newcount\Comments  
\definecolor{darkgreen}{rgb}{0,0.5,0}
\definecolor{darkred}{rgb}{0.7,0,0}
\definecolor{teal}{rgb}{0.1,0.6,0.7}
\definecolor{blue}{rgb}{0.0,0.1,0.9}
\definecolor{orange}{rgb}{1.,0.7,0.0}
\definecolor{lightblue}{rgb}{0.70, 0.80, 0.89}
\definecolor{violet}{rgb}{0.50, 0.16, 0.88}
\newcommand{\kibitz}[2]{\ifnum\Comments=1{{\textcolor{#1}{\textsf{\footnotesize [#2]}}}}\fi}

\title{Entity Linking and Discovery via \\Arborescence-based Supervised Clustering
}

\author{Dhruv Agarwal, Rico Angell, Nicholas Monath, Andrew McCallum \\
        College of Information and Computer Sciences \\
        University of Massachusetts Amherst \\ 
        \texttt{\{dagarwal,rangell,nmonath,mccallum\}@cs.umass.edu}}

\begin{document}
\maketitle

\begin{abstract}
Previous work has shown promising results in performing entity linking by measuring not only the affinities between mentions and entities but also those amongst mentions. In this paper, we present novel training and inference procedures that fully utilize 
mention-to-mention affinities by building minimum arborescences (i.e., directed  spanning trees) over mentions and entities across documents in order to make linking
decisions. We also show that this method gracefully extends to entity discovery, enabling the clustering of mentions that do not have an associated entity in the knowledge-base. We evaluate our approach on the Zero-Shot Entity Linking dataset and MedMentions, the largest publicly available biomedical dataset, and show significant improvements in performance for both entity linking and discovery compared to identically parameterized models. We further show significant efficiency improvements with only a small loss in accuracy over previous work, which use more computationally expensive models.
\end{abstract}
\section{Introduction}

Entities are often mentioned ambiguously in natural language corpora, 
such as biomedical research papers \cite{leaman2016taggerone,sung-etal-2020-biomedical}
, news \cite{milne2008learning,hoffart-etal-2011-robust}, and web page text
\cite{gabrilovich2013facc1,lazic2015plato}. 
Resolving the ambiguity of these entity mentions requires 
either linking each mention to a knowledge-base (KB)
or, if there is no suitable KB entry, adding a 
new entity to the KB. The latter task, \emph{entity discovery},
is often done by discovering coreference relationships 
among mentions and providing each coreferent mention  
an identifier that represents the newly added entity 
\cite{mcnamee2009overview, radford2011naive}. 
Linking and discovery are important 
for question answering \cite{das2019multi} and
building KBs \cite{ling-etal-2015-design}
or semantic indexes \cite{leaman2016taggerone}.

Entity linking is particularly challenging 
in zero-shot settings, where 
not every entity has labeled training data \cite{lin-etal-2017-list,logeswaran-etal-2019-zero}.  
In such settings, we rely on entity descriptions, 
types, and aliases 
to form entity representations, which are used for linking predictions.

\begin{figure}
    \centering
    \includegraphics[width=\columnwidth]{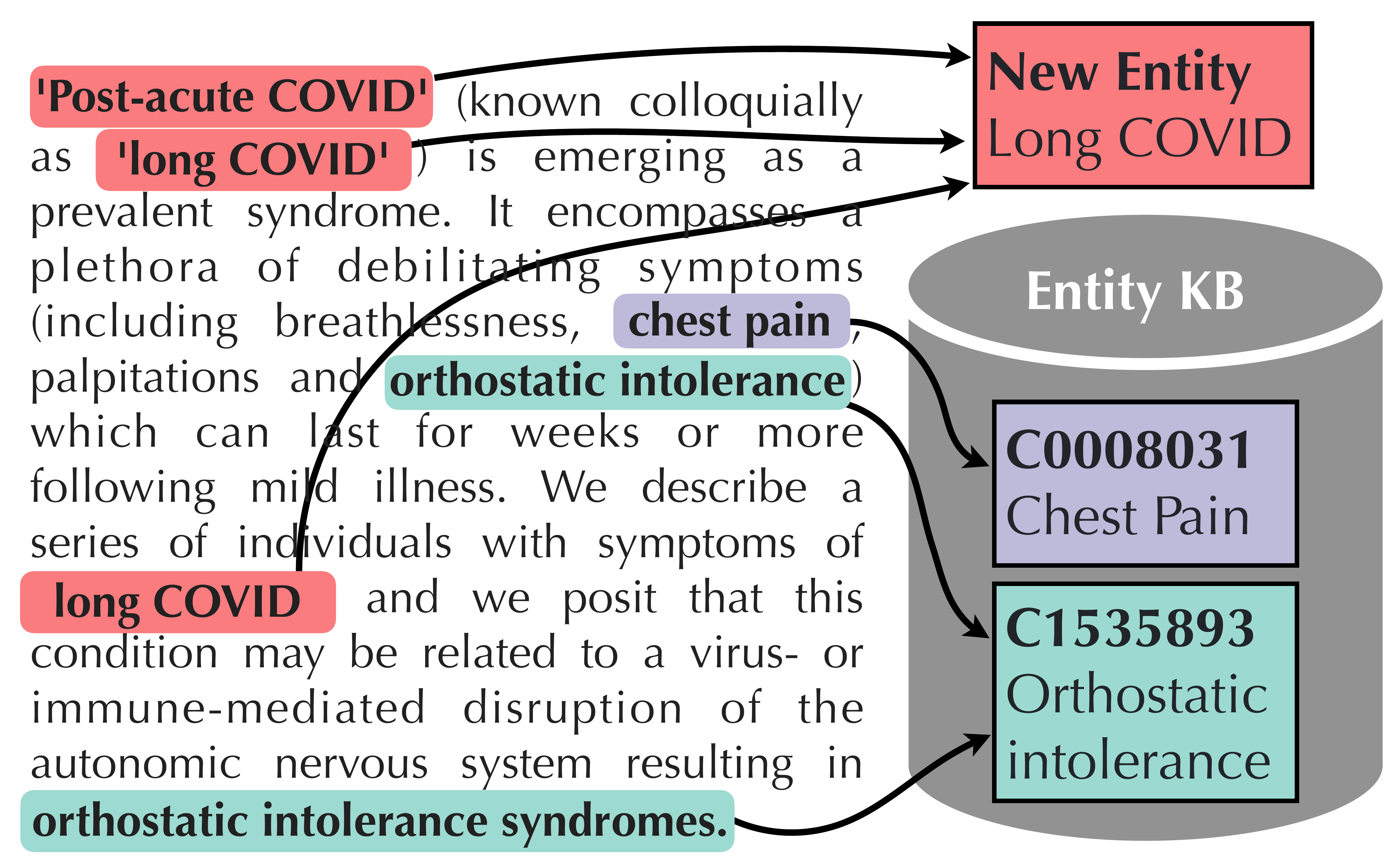}
    \caption{\textbf{Linking \& Discovery}. We consider the task of both entity linking and discovery (also known as NIL clustering), in which ambiguous mentions are either ground to existing KB entities or form new ones.}
    \label{fig:entity_discovery_task}
\end{figure}

Entity linking and coreference are closely related, with linking decisions implying coreference 
relationships amongst mentions \cite{dutta-weikum-2015-c3el}.
Recent work
\cite{angell2021clusteringbased} has demonstrated how
a clustering-based approach for entity linking prediction
can share linking decisions by
predicting coreference between mentions. 
This approach, however, uses 
cross-encoder based transformer models \cite{devlin-etal-2019-bert}, which are prohibitively expensive
when it comes to running in cross-document settings --- the number of input sequences passed through the transformer encoder scales quadratically with the number of mentions. This limits the approach to only consider pairs of mentions within the same document when computing mention-to-mention affinities.

Inspired by
this recent work on clustering-based inference, 
we present a new 
model for clustering that uses a graph-based approach
of modeling directed
nearest-neighbor relationships among mentions
in an \emph{arboresence} (directed minimum spanning tree).
We
demonstrate how this approach uses a bi-encoder \cite{devlin-etal-2019-bert} in order to be significantly more efficient.
To complement this, we propose a supervised clustering training objective
motivated by our inference procedure.
We further show how our model can be used as a unified approach for both linking entities as well as discovering them. 

We evaluate our approach on 
two entity linking datasets, both of which 
are in the zero-shot 
domain. We compare the performance of our training procedure against two standard bi-encoder training procedures using both independent inference (linking each mention individually) and clustering-based inference. We find that our training improves performance by 13.1 percentage points on MedMentions and 11.1 \& 0.6 points on the two training variants compared on ZeShEL. 
In addition, we run experiments for entity discovery by removing a fraction of entities from the KB and demonstrate that our training and inference procedures are better suited for this task. Compared to cross-encoder based approaches, our
bi-encoder approach is more that two times more efficient with only a small loss in accuracy. 

\section{Problem Definition}

\begin{figure*}[t!]
    \centering
    \includegraphics[scale=0.34, trim=0 0 0 0, clip]{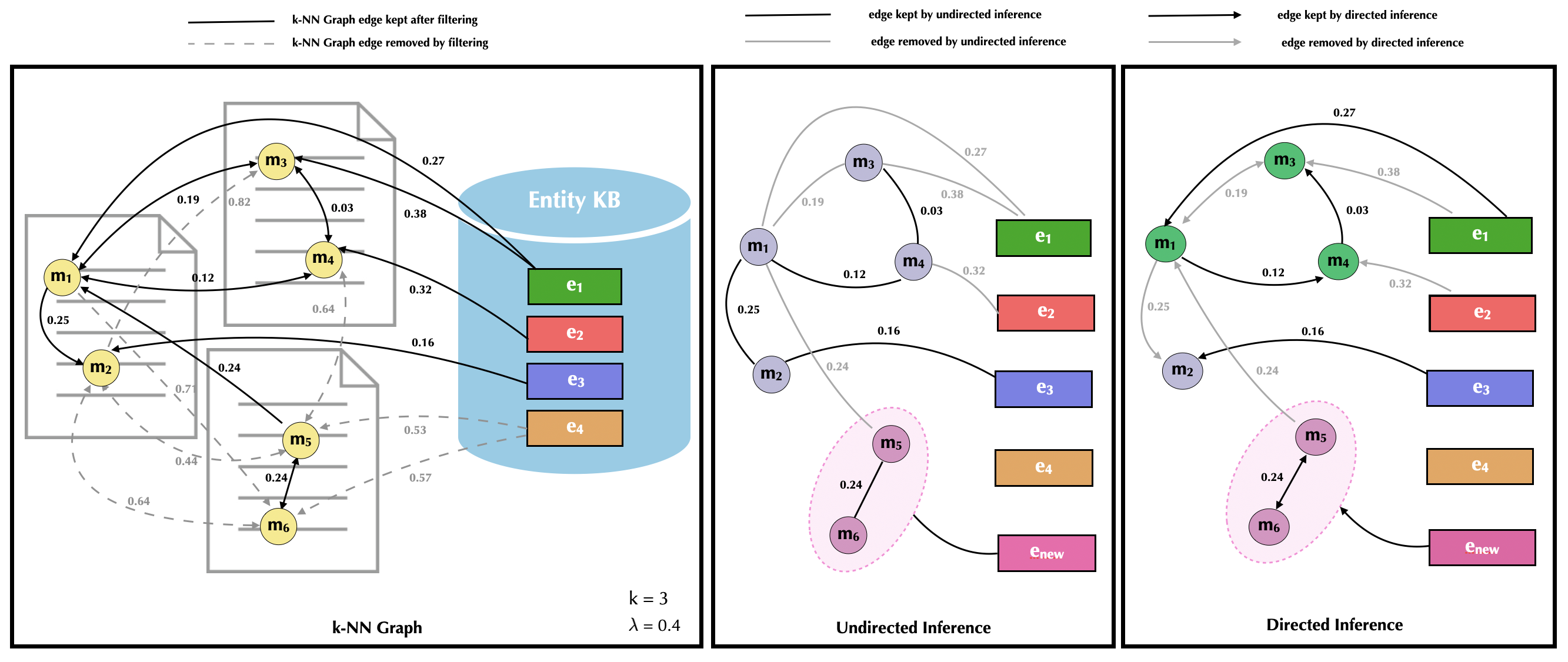}
    \caption{\textbf{Inference for Entity Linking and Discovery}. Mentions are represented by circles and entities by rectangles. The left sub-figure shows the $k$-nearest neighbor graph over all mentions across all documents and all entities in the knowledge-base. Dashed edges are edges that were dropped since they have edge weights above the threshold $\lambda=0.4$ and solid edges represent the resulting graph, which is input to the inference algorithms. The center sub-figure shows the result of undirected inference and the right figure shows the result of directed inference. In both sub-figures, the dark edges are the edges kept and the light edges are those removed by the inference procedure. Both inference procedures result in one cluster that does not contain an entity,  predicting this cluster of mentions as a new discovered KB entity.}
    \label{fig:inference}
\end{figure*}

Each document $\doc$ of a corpus $\docs$
contains a set of entity mention spans
$\mentions^\doc = \{\mention^\doc_1,\mention^\doc_2,\dots,\mention^\doc_N\}$. 
All mentions in the corpus are given by $\mentions = \bigcup_{\doc \in \docs} \mentions^\doc$.
Following \cite{logeswaran-etal-2019-zero,angell2021clusteringbased},
we assume that these mentions
are pre-identified spans of text.

\paragraph{Entity Linking} We first consider the task
of entity linking in which we are provided a knowledge-base 
 of entities $\KB$ and our task is to predict an entity $\entity^\doc_i \in \KB$
 for each mention $\mention^\doc_i$. We use ${\entity^\star}^\doc_i$ to
 refer to the ground truth entity label for $\mention^\doc_i$.
 
 \paragraph{Zero-Shot Linking} The zero-shot task refers to the setting where there are entities in the knowledge base that do not have any labeled training data. Linking decisions must, instead, rely on provided
 information for entities such as a description, aliases, and/or the 
 entity type.

 \paragraph{Linking + Discovery} We also
 consider a setting in which the complete knowledge-base 
 of entities may not be known in advance and new entities must
 be \emph{discovered}. For this task, 
 we assign every entity mention $\mention^\doc_i$ a cluster/coreference 
 label $\cluster^\doc_i \in \clusters$ that is independent of the entity labels in the KB, i.e. $\clusters \cap \KB = \emptyset$. 
 In this setting, a mention $\mention^\doc_i$ may be assigned $\entity^\doc_i=\NIL$, indicating
 that $\mention^\doc_i$ refers to an entity not present in the knowledge-base. 
 Non-$\NIL$ assignment decisions imply coreference between mentions.

 It is important to note the distinctions between zero-shot linking and discovery.
 In the zero-shot setting, entities are \textit{known} in the KB ahead of time,
 but there is no training data. In the discovery setting, we do not 
 know all the entities \textit{a priori}, and our performance, then, is evaluated in terms of the correctness of clustering.

\section{Linking \& Discovering Entities with Graph-Based Clustering}

In this section, we describe our proposed approach 
for making entity linking and discovery decisions.
We define a distance measure between 
mentions and entities in terms of a directed graph,
where nodes refer to both mentions as well as entities, and 
the edges are inferred by the model.
Clusters are then constructed from this graph such that
no cluster contains more than one entity. Linking decisions are made by assigning each mention within a cluster to the entity node
present within the same cluster. In this way, clusters represent 
new (discovered) entities if a set of mentions is clustered
without an entity node.

 Our approach generalizes the clustering-based approach
of \citet{angell2021clusteringbased} by (1) 
proposing a clustering approach on \emph{directed} graphs 
(2) using all mentions in the corpus rather than operating in a within-document setting.

There are four main components to our approach: (1) graph-based 
dissimilarity measure, (2) models to provide edge weights, (3) approach for inferring latent graph, and (4) building constrained clusters.

\paragraph{Graph-based Dissimilarity} Let $\graph$ be a graph with nodes $\vertices = \mentions \cup \KB$ and 
directed edges $\edges \subset \vertices \times \vertices$. Each
edge $(x,y)$ of the graph has an associated weight $w_{x,y}$. 
We define a dissimilarity function $f$ between two nodes $u,v \in \vertices$ to be
the weight of the minimax path between the nodes, i.e. 
{{\begin{equation}
\small
    f(u, v) = \begin{cases} \min\limits_{p \in u \rightsquigarrow v} \max\limits_{(x,y) \in p} w_{x,y}, & \text{if } \mathop{\rm connected}(u,v) \\
    \;\;\infty, & \text{otherwise}
    \end{cases}
\end{equation}}}

\noindent\!\!where $\mathop{\rm connected(u, v)}$ is true if there exists a directed path from node $u$ to $v$ in $G$, and $u \rightsquigarrow v$ is the set of all paths between $u$ and $v$. In words, the dissimilarity between $u$ and $v$ is the minimum of the highest weight edges in all paths between the two nodes, and this is often referred to as the "bottleneck edge". This measure has the property of emitting 
low dissimilarities between nodes even when the direct edge weight $w_{u,v}$ is high by connecting them through a chain of 
low-weight edges providing an inductive bias well-suited for coreference, i.e. not all pairs of points in a cluster are nearby, see Figure~\ref{fig:inference} for an example. This inductive bias is not achieved if we sum edge weights and simply find the minimum path.

\paragraph{Edge Weights}
With this definition of dissimilarity, we now define how edge weights 
are calculated. 
We use two models: 
a mention-pair affinity model, $\mentionmention: \mentions \times \mentions \rightarrow \RR$,
and a mention-entity affinity model, $\mentionentity: \KB \times \mentions \rightarrow \RR$.
An edge between two mentions $\mention_i$ and $\mention_j$ has weight:
\begin{equation}
\begin{aligned}
    w_{\mention_i,\mention_j} &= -\, \mentionmention(\mention_i,\mention_j),
\end{aligned}
\end{equation}
and the weight of the edge from entity $\entity$ to $\mention_i$ is:
\begin{equation}
\begin{aligned}
    w_{\entity, \mention_i} &= -\, \mentionentity(\entity,\mention_i)
\end{aligned}
\end{equation}
Each of $\mentionmention(\cdot,\cdot)$ and $\mentionentity(\cdot,\cdot)$ are parameterized by bi-encoder transformer models \cite{gillick2019learning, humeau2019poly}. We train two independently parameterized transformer encoder models: one for mentions, $\mentionencoder$, and one for entities, $\entityencoder$. The affinity models
are simply the inner products of the associated encoded representations:
\begin{equation}
\begin{aligned}
    \mentionmention(\mention_i, \mention_j) &=  \mentionencoder(\mention_i)^T\mentionencoder(\mention_j) \\
    \mentionentity(\entity, \mention_i) &=  \entityencoder(\entity)^T\mentionencoder(\mention_i)
\end{aligned}
\end{equation}
\noindent For the mention encoder, $\mentionencoder$, the input to the 
transformer is the mention's surrounding context with the mention 
span marked by special tokens \texttt{[START]} and \texttt{[END]}:
\begin{align*}
    &\texttt{[CLS]} c_\text{left} \texttt{[START]} m_i \texttt{[END]} c_\text{right} \texttt{[SEP]}
\end{align*}
where $c_\text{left}$ and $c_\text{right}$ are the left and right contexts of the mention $m_i$ in the document. For the entity encoder, $\entityencoder$, the transformer takes as input the title and description of the entity: 
\begin{equation*}
\texttt{[CLS]} e_\text{title} \texttt{[TITLE]}e_\text{desc}\texttt{[SEP]}
\end{equation*}
In this input, $e_\text{desc}$ is the token sequence corresponding to the description of the entity, which could include natural language text related to the entity, such as a "wiki" entry or a list of synonym representations of the entity, or any other available features useful in forming an entity representation.

\paragraph{Building the Graph} The structure of the graph $G$ impacts the 
dissimilarity function by changing the paths between pairs of nodes in addition to changing which pairs of nodes are connected.
We advocate for a simple, deterministic approach to construct this graph. For each mention $m$, construct $E_{m}$ by (1) adding edges from $m$'s $k$-nearest neighbor mentions in $\mentions$ to $m$, and (2) adding an edge from $m$'s nearest entity to $m$:
\begin{equation}
\begin{aligned}
E_m &= \left\{ (u, m) \ \Big| \  u \in  \mathop{\rm argmink}_{m' \ \in \ \mentions} w_{m',m} \right. \\
& \qquad \qquad \quad\; \left. \lor \;  u = \argmin_{\entity \ \in \ \KB} w_{\entity,m} \right\}
\end{aligned}
\end{equation}
The complete collection of edges $E$ in $G$ is given by $E(G) = \bigcup_{m \in \mentions} E_{m}$. There are other ways that one could conceivably pick the pairs of mentions to be connected in the graph. For example, one could use the minimum spanning tree over the mentions. This approach, however, has several drawbacks: (1) the directionality of nearest neighbor relationships is ignored leading to added noise in the graph, and (2) the resultant graph includes edges that clearly cross cluster boundaries due to this approach forcing all pairs of mentions to be connected.

\paragraph{Forming Clusters \& Making Predictions} The graph $G$ is input
to a constrained clustering problem that partitions $G$ into disjoint clusters $\clusters = \{C_1,\dots,C_M\}$ such 
that each cluster contains at most one entity. There are three constraints that every $C \in \clusters$ must satisfy: 
\begin{equation*}
    \begin{aligned}
        \small
        &\textrm{  i. } \; |C \cap \KB| \leq 1,  \qquad \quad \;\, 
        \\
        &\textrm{ ii. } \; \forall u,v \in C, \; \mathop{\rm connected}(u,v) \implies f(u,v) \leq \lambda ,
        \\
        &\textrm{iii. } \; \forall u,v \in C, \; \mathop{\rm connected}(u,v) \vee \mathop{\rm connected}(v, u)
    \end{aligned}
\end{equation*}
where $\lambda$ is a specified hyperparameter representing the dissimilarity threshold. These constraints ensure that (i) there is at most one entity in each cluster, (ii) if $u$ is reachable from $v$ then every edge in the path from $v$ to $u$ has a weight $\leq \lambda$, and (iii) each node in the cluster has a path connecting it with every other node in the cluster. We solve this constrained clustering problem, i.e., partition graph $G$, 
using a process similar to \citet{angell2021clusteringbased}.

Specifically, we first remove all edges in graph $G$ with weight greater than $\lambda$. We then evaluate each edge $(u,v) \in E$ in descending order of dissimilarity and check if its presence violates any of the three constraints defined above, removing the edge from $E$ if it does. If not, we evaluate whether there is an entity in the connected component of node $u$, i.e. $|C_u \cap \KB| = 1$. We infer that $C_u$ refers to a new entity not present in the KB if an entity is not found. If, however, $|C_u \cap \KB| = 1$, we temporarily drop edge $(u,v)$ and check whether $v$ can still be reached by an entity node. If reachable, we permanently drop $(u,v)$, maintaining the validity of constraint (i) as well as our minimax dissimilarity function $f(\cdot,\cdot)$. If an entity cannot reach $v$, we retain edge $(u,v)$, preserving the connectivity of the cluster, and iterate further. Our predicted clusters are the resultant connected components in the partitioned graph $G$. We refer to this procedure as \emph{directed inference} when we respect the direction of the edges and \emph{undirected inference} when we disregard the direction (see Figure~\ref{fig:inference}).

To make linking decisions for each mention $m^d_i$, we assign the ID of the entity present in the mention's cluster as the linking label (or $\NIL$ if there is no entity in the cluster). Let $\clusters(m^d_i)$ be the predicted cluster of mention $m^d_i$, then:
\begin{equation}
    e^d_i = \begin{cases} \clusters(m^d_i) \cap \KB, & \text{if } |\clusters(m^d_i) \cap \KB| = 1 \\
    \NIL, & \text{ otherwise}
    \end{cases}.
\end{equation}
Furthermore, the clusters we predict for in the entity discovery setting are exactly $\clusters$.

\paragraph{Cluster Spanning Arborescence} 
For every cluster with an entity node, the edge structure is a directed analogue of the minimum spanning tree where there is a directed path from the entity node to every other node in the cluster. This structure is often referred to as the \emph{minimum spanning arborescence}, thus lending its name to our method, i.e. \ours linking and discovery.

\section{Training}
In this section, we present our approach for training
the affinity models, $\mentionmention(\cdot, \cdot)$ and $\mentionentity (\cdot, \cdot)$,
and their associated encoders, $\mentionencoder$ and $\entityencoder$.
Our objective is to optimize the dissimilarity function $f(\cdot, \cdot)$
such that the clustering procedure infers a set of clusters that each contains exactly one entity, and 
every mention is assigned to the cluster containing its ground truth entity. 
We optimize $f(\cdot,\cdot)$ 
using mini-batch gradient descent methods. 
To achieve this, we sequentially build mini-batches of mentions 
$\minibatch \subset \mentions$ over the training data, where each $\mention_i \in \minibatch$
has ground truth entity $\entity^\star_i$. We then build a
graph $\graph_\minibatch$, where the nodes are all $m_i \in \minibatch$, all mentions coreferent to $m_i \in \minibatch$, and the set of ground truth entities for each $m_i \in \minibatch$.

For each $\mention_i$, we build a set of edges,
\begin{equation}
\begin{aligned}
E_{m_i} &=\left\{ (e^\star_i, \mention_\ell) \ \Big| \ m_\ell \in \mentions_{e^\star_i} \right\} \\
& \qquad \cup \left\{(\mention_\ell, \mention_p) \ \Big| \ \mention_\ell,\mention_p \in \mentions_{e^\star_i} \right\}
\end{aligned}
\end{equation}
The complete set of edges in graph $\graph_\minibatch$ for a 
mini-batch $B$ is then given 
by $E(\graph_\minibatch) = \bigcup_{\mention_i \in \minibatch} E_{\mention_i}$. 
Observe that the resultant edges ensure that each connected component 
contains exactly one entity (namely, the ground truth entity for the mentions in that component). We then sparsify $\graph_\minibatch$ by 
computing a partitioned target graph $\graph^\star_\minibatch = \{E^\star_{\mention_i} \ | \ \mention_i \in \minibatch\} $ using the inference procedure defined in the previous section, setting $\lambda = \infty$. After this sparsification, $\graph^\star_\minibatch$ becomes a disjoint set of minimum spanning arborescences 
rooted at the entity nodes.
We use $E^\star_{\mention_i}$ to optimize
the parametric encoder models. Note that each mention node in a target edge set $E^\star_{\mention_i}$ has only one 
incoming edge originating from either an entity or a mention, and the selection of $E^\star_{\mention_i}$ was done in a way to minimize $f(\cdot, \cdot)$ between mentions and entities with the same label (maximize opposite labels) on the subgraph of the mini-batch.

Akin to the graph embedding objectives used by \citet{nickel2018learning} 
and others, we construct our objective by sampling negative edges.
For each  mention $\mention_i \in \minibatch$, the set of negative edges $\texttt{N}(\mention_i)$ is the $k/2$ lowest-weight incoming edges from $\mathcal{E}\setminus \{e^\star_i\}$ and the $k/2$ lowest-weight incoming edges from $\mathcal{M}\setminus \mathcal{M}_{e^\star_i}$, where $k$ is a specified hyperparameter.
Take $\Gamma(\mention_i) = \{u \ | \ (u, m_i) \in E^*_{m_i}\} \cup \{u \ | \ (u, m_i) \in \texttt{N}(\mention_i)\}$ to be the set of all neighbors with an outgoing edge to $\mention_i$ in the training graph. Let $\II_{u, m_i}$ be the indicator variable such that $\II_{u, m_i} = 1$ if $(u, m_i) \in E^*_{m_i}$ and $\II_{u, m_i} = 0$ otherwise. Our loss function with respect to each mention $m_i \in B$ is as follows:
\begin{align}
    \mathcal{L}(m_i) &= \!\!\!\sum_{u \in \Gamma(m_i)} \!\!\!\Big( \II_{u, m_i} \log(\sigma_u(w_{u, m_i})) \\
     & \quad + (1 - \II_{u, m_i})\log(1 - \sigma_u(w_{u, m_i})) \Big), \nonumber
\end{align}
where $\sigma(\cdot)$ is the softmax function over all edges in $\Gamma(m_i) \times \{m_i\}$. The loss for the entire batch $B$ is the mean of losses over all mentions in $B$. Optimizing this loss function requires simultaneously increasing the likelihood of the positive edges and decreasing the likelihood of the negative edges. This objective and training routine are inspired by the supervised single-linkage clustering proposed by \citet{pmlr-v97-yadav19a}, but differs in the choice of loss function and selection of negative examples. We also experimented with the standard cross-entropy loss, but found its performance subpar.
\begin{table*}[]
    \centering
    \resizebox{\textwidth}{!}{\begin{tabular}{llcccc|cc}
    \toprule
    && \multicolumn{4}{c|}{\bf MedMentions} & \multicolumn{2}{c}{\bf \zeshel} \\
    \bf Training & \bf Inference & &  \bf Overall & \multicolumn{2}{c|}{\bf Acc. on } & & \bf Overall  \\
    && \bf k &  \bf Acc.    &\bf Seen & \bf Unseen  & \bf k &    \bf Acc. \\
    \midrule
    
    & Independent$^{\dagger}$ & - & 58.7 & 61.4 & 49.2 & - & 39.3 \\ 
    \inBatch & k-NN Graph \small{\textsc{(undirected)}} & 1 & 59.1 & 61.9 & 49.4 & 1 & 38.5 \\
    & k-NN Graph \small{\textsc{(directed)}} & 1 & 59.1 & 62.0 & 49.2 & 1 & 38.6 \\
    \midrule[0.01pt]
    
    & Independent$^{\dagger}$ & - & 56.9 & 64.0 & 31.9 & - & 49.8 \\ 
    \knnNegs & k-NN Graph \small{\textsc{(undirected)}} & 1 & 51.0 & 58.0 & 26.4 & 1 & 40.1 \\
    & k-NN Graph \small{\textsc{(directed)}} & 1 & 52.8 & 60.5 & 25.9 & 1 & 40.8 \\
    \midrule[0.01pt]
    
    & Independent$^{\dagger}$ & - & \bf 72.3 & 77.5 & 54.2 & - & 50.3 \\ 
    \ours \tiny{(Ours)} & k-NN Graph (\small{\textsc{undirected}}) & 1 & \bf 72.3 & \bf 77.6 & \bf 54.3 & 1 & 50.3 \\
    & k-NN Graph \small{\textsc{(directed)}} & 2 & 72.2 & 77.5 & 53.8 & 1 & \bf 50.4 \\

    \bottomrule
    \end{tabular}}
    \caption{\textbf{Bi-Encoder Linking Results: Accuracy} ($^{\dagger}$Predictions based on mention-to-entity similarity only)}
    \label{tab:biencoder_accuarcy}
\end{table*}
\begin{table}[]
\footnotesize{
    \centering
    \begin{tabular}{l@{}c@{}c}
    \toprule
    \bf Training  &  \multicolumn{2}{c}{\bf Recall@64} \\
    & \bf MedMentions \hspace{1mm} & \bf \zeshel \\
    \midrule[0.01pt]
    \inBatch & 87.69 & 84.04\\
    \knnNegs & 85.84 & 84.77\\
    \ours \tiny{(Ours)} & \bf 95.62 & \bf 85.11\\    
    \bottomrule
    \end{tabular}
    \caption{\textbf{Bi-Encoder Linking Results: Recall@64}}
    \label{tab:biencoder_recall}
    }
\end{table}

\section{Experiments}

We perform entity linking and discovery experiments
using two datasets that require generalization
to unseen entities at test time (Table~\ref{tab:dataset_stats}). 

We analyze the improvements using 
our proposed approach as compared to
similarly parameterized bi-encoder methods. 
We provide an analysis of the performance of each 
component of our approach, comparing our proposed
graph clustering and training objective to 
sensible baselines, and show that our model has much higher accuracy while being as efficient as these baselines. We further compare
our approach to a state-of-the-art cross-encoder
method and show that our approach produces
comparable accuracy results while being much more
efficient.

\subsection{Datasets}
\label{sec:dataset}

\paragraph{MedMentions \cite{mohan2019medmentions}\footnote{\url{https://github.com/chanzuckerberg/MedMentions}}} is a collection of titles and abstractions of 
bio-medical research papers. The KB
that is used for this dataset is the 2017AA full-version of UMLS. 
The validation and test sets contain both entities that
are present in the training set as well as entities that
are zero-shot (never seen at training time). We use the author-recommended ST21pv subset. 

\paragraph{\zeshel \cite{logeswaran-etal-2019-zero}} 
is a collection of Fandom Wikias. The Wikias 
are divided into train / dev / test splits, with no set overlapping in entities. In this 
way, all entities that appear at validation and test time
are not seen during training.

\paragraph{Entity Discovery} For this setting, we modify each dataset by randomly sampling 10\% of the entities 
in the dev and test partitions, and remove them from the 
knowledge-base. These removals are the new entities that
we attempt to discover in this task. 
To our knowledge, other ED systems include \cite{andrews2014robust}, \cite{dutta2015c3el}, \cite{pershina2015personalized}. However, incorporating BERT-based encoder models in these approaches is not straightforward, therefore a comparison was outside the scope of this work.

\subsection{Methods Compared}

We analyze our proposed approach through comparisons
with various sensible alternatives 
for graph clustering and 
training. 

\paragraph{Training Objectives} We compare
the proposed graph-based training objective, 
which directly trains both the mention-mention
similarity function $\mentionmention(\cdot,\cdot)$
and the mention-entity similarity function $\mentionentity(\cdot,\cdot)$
to baselines, which only explicitly train $\mentionentity(\cdot,\cdot)$
(and rely on the structure of $\mentionmention(\cdot,\cdot)$ sharing 
representions with $\mentionentity(\cdot,\cdot)$ to provide meaningful
mention-mention similarities). We compare to two baselines: 
(1) training $\mentionentity(\cdot,\cdot)$ with random negatives (\textsc{\inbatch}) and (2) training $\mentionentity(\cdot,\cdot)$ with hard negatives (\textsc{\hardnegatives}).

\paragraph{Linking Procedures} We compare our proposed
clustering-based inference procedure to a state-of-the-art independent procedure (\independent), which directly links each mention to its nearest
entity. This model was used by \citet{wu2019zero} to generate candidates
for a cross-encoder model trained on \zeshel. We further
compare our approach to using \emph{undirected} rather than \emph{directed} edges 
in the graph clustering step. 

\paragraph{Cross-Encoder Models / SOTA} To measure how much we pay in accuracy to use
these more efficient bi-encoder models vis-\`a-vis cross-encoder models, we
compare our performance to the state-of-the-art cross-encoder based model 
that uses clustering-based inference as that is most directly comparable to our method.  \citet{angell2021clusteringbased} train two cross-encoder models, one for mention-mention affinities and one for mention-entity affinities, and use an inference procedure equivalent to our undirected clustering-based inference.
The cross-encoder procedure they use necessitates limiting the potential mention-mention and mention-entity edges in the graph used for clustering since it is intractable to consider all pairs, while our method requires no such restriction.

\begin{table*}[]
    \centering
    \resizebox{\textwidth}{!}{\begin{tabular}{llcccc|cccc}
    \toprule
    && \multicolumn{4}{c|}{\bf MedMentions} & \multicolumn{4}{c}{\bf \zeshel} \\
    \bf Training & \bf Inference & \bf k & \bf NMI &\bf ARI & \bf $\frac{\text{NMI+ARI}}{\text{2}}$ & \bf k & \bf NMI &\bf ARI & \bf $\frac{\text{NMI+ARI}}{\text{2}}$ \\
    \midrule
    
    & No Entities$^{\dagger}$ & - & 0.93 & 0.37 & 0.65 & - & \bf0.98 & 0.31 & 0.65\\ 
    \inBatch & k-NN Graph \small{\textsc{(undirected)}} & 4 & \bf0.95 & 0.58 & 0.76 & 1 & 0.96 & 0.25 & 0.61 \\
    & k-NN Graph \small{\textsc{(directed)}} & 8 & \bf0.95 & 0.58 & 0.76 & 8 & 0.96 & 0.17 & 0.57 \\
        \midrule[0.01pt]

    & No Entities$^{\dagger}$ & - & 0.93 & 0.49 & 0.71 & - & \bf0.98 & 0.29 & 0.64\\ 
    \knnNegs & k-NN Graph \small{\textsc{(undirected)}} & 1 & 0.93 & 0.56 & 0.75 & 1 & \bf0.98 & 0.31 & 0.64 \\
    & k-NN Graph \small{\textsc{(directed)}} & 1 & 0.93 & 0.57 & 0.75 & 1 & \bf0.98 & 0.31 & 0.64 \\
        \midrule[0.01pt]

    & No Entities$^{\dagger}$ & - & 0.94 & 0.51 & 0.72 & - & \bf0.98 & 0.34 & 0.66\\ 
    \ours \tiny{(Ours)}& k-NN Graph \small{\textsc{(undirected)}} & 8 & \bf0.95 & \bf0.64 & \bf0.79 & 1 & 0.97 & 0.27 & 0.62 \\
    & k-NN Graph \small{\textsc{(directed)}} & 8 & \bf0.95 & \bf0.64 & \bf0.79 & 1 & \bf0.98 & \bf0.36 & \bf0.67 \\
    
    \bottomrule
    \end{tabular}}
    \caption{\textbf{Bi-Encoder Discovery Results.} ($^{\dagger}$Predictions based on mention-to-mention similarity only)}
    \label{tab:biencoder_discovery}
\end{table*}
{\tiny { \begin{table}[]
    \centering
    \footnotesize{
    \begin{tabular}{cl cc }
    \toprule
    & & \bf MedMentions & \bf \zeshel \\
     \midrule
    \multirow{3}{*}{$|\mentions|$} & Train & 120K & 49K  \\
    & Dev & 40K & 10K\\
    & Test & 40K & 10K \\
    \midrule[0.01pt]
    \multirow{3}{*}{$|\KB|$} & Train & 19K & 26K \\
    & Dev & 9K & 7K \\
    & Test & 8K & 7K \\
    \midrule[0.01pt]
    \multirow{2}{*}{$ |\KB \setminus \KB_{\tiny{\text{Train}}}|$} 
    & Dev & 4K & 7K\\
    & Test & 4K & 7K\\
    \bottomrule
    \end{tabular}
    \caption{\textbf{Dataset Statistics}. $|\mentions|$ is the number of mentions. $|\KB|$ is the number of unique entities in the labeled partition, not the total KB size. $|\KB \setminus \KB_\text{Train}|$ is the number of \emph{zero-shot} entities. The total KB size of MedMentions and \zeshel are  2M and 500K respectively.}
    \label{tab:dataset_stats}}
\end{table}
}}

\subsection{Linking Results}
We report the linking accuracy on MedMentions and \zeshel
in Table \ref{tab:biencoder_accuarcy}. We refer to our training procedure as \ours, and show comparisons in performance of our model with \inBatch and 
\knnNegs, as described in the previous section. We use three inference procedures, including our clustering-based approach in both \textsc{Directed} and \textsc{Undirected} modes. We report test set accuracy at $k$ that performed best on the validation set.

\paragraph{MedMentions} We 
follow previous work in not supplying gold entity type information during inference. This makes the task significantly more challenging. In 
addition to reporting the overall linking accuracy, we measure the accuracy on \textit{Seen} and 
\textit{Unseen} subsets, which correspond to the ground truth entities that were seen and unseen (i.e., zero-shot) at training, respectively.

In Table~\ref{tab:biencoder_accuarcy}, we observe that regardless of the chosen inference, the \ours model significantly outperforms the other models by at least 13.1 points in terms of overall accuracy, 13.5 points on \textit{Seen}, and 4.4 points on \textit{Unseen} mentions. We further see that \textsc{Undirected} clustering marginally outperforms the \textsc{Directed} variant in the three categories by 0.1, 0.1, and 0.5 points, respectively. In Table \ref{tab:biencoder_recall}, we report Recall@64, defined as the accuracy of predicting a gold entity from any of the top-64 predictions for a mention. We see that \ours achieves 7.93 and 9.78 points of improvement over In-Batch and k-NN models. 

These improvements in accuracy indicate that our training procedure enables the model to learn much better representations of mentions and entities than previous approaches.

We also compare the efficiency / accuracy trade-offs
between our models and \citet{angell2021clusteringbased}.
Table~\ref{tab:ce_comparison} shows the training time, inference time, and accuracy for all training and inference procedures considered on MedMentions. In less than half of the training and inference time, without any external information or preprocessing, and a much weaker architecture, our method is able to achieve accuracy within 2 points of the SOTA cross-encoder model. 
This highlights the effectiveness of our approach in producing a highly accurate, yet efficient, model. 

\vspace{1mm}

\noindent\textbf{\zeshel} On this dataset, we evaluate assuming gold entity type information is provided at both training and inference time. Since entities in the validation and test sets are from domains different than those in the test set, our evaluation is fully zero-shot. In Table \ref{tab:biencoder_recall}, we show that the \ours model achieves better Recall@64 by 1.07 points and 0.34 points over \inbatch and k-NN, respectively. In Table \ref{tab:biencoder_accuarcy}, we present accuracy results, which prior work using bi-encoder models, to the best of our knowledge, does not report. We see that the \ours model is more accurate than the \inbatch and k-NN models by 11.1 and 0.6 points of accuracy, respectively.

\begin{table*}[]
    \centering
    \resizebox{\textwidth}{!}{\begin{tabular}{lc | lc |c}
    \toprule
    \bf Training & \bf Time (hrs) & \bf Inference & \bf Time (hrs) & \bf Accuracy\\
    \midrule
    
    & & Independent$^{\dagger}$ & 1.5 & 58.7\\ 
    \inBatch & 3.4 & k-NN Graph \small{\textsc{(undirected)}} & 1.7 & 59.1\\
    & & k-NN Graph \small{\textsc{(directed)}} & 1.7 & 59.1\\
        \midrule[0.01pt]

    & & Independent$^{\dagger}$ & 1.5 & 56.9\\ 
    \knnNegs & 18.5 & k-NN Graph \small{\textsc{(undirected)}} & 1.7 & 51.0\\
    & & k-NN Graph \small{\textsc{(directed)}} & 1.7 & 52.8\\
            \midrule[0.01pt]

    & & Independent$^{\dagger}$ & 1.5 & 72.3\\ 
    \ours \tiny{(Ours)} & 32.1 & k-NN Graph \small{\textsc{(undirected)}} & 1.6 & 72.3\\
    & & k-NN Graph \small{\textsc{(directed)}} & 1.6 & 72.2\\
                \midrule[0.01pt]

    \clusterbased\tiny{~\cite{angell2021clusteringbased}} & 72.0 & k-NN Graph \small{\textsc{(undirected)}} & 4.0 & 74.1\\
    
    \bottomrule
    \end{tabular}}
    \caption{\textbf{Efficiency-Accuracy Trade-Off} ($^{\dagger}$Predictions based on entity-to-mention affinity only)}
    \label{tab:ce_comparison}
\end{table*}

\subsection{Entity Discovery Results}

In this setting, 
we evaluate the predicted clusters of our proposed approach
on a modified test set for MedMentions and \zeshel by
removing 10\% of the entities appearing in the test data from the 
knowledge-base and training new models with the held-out entities removed during training. We report performance
in terms of two frequently-used clustering 
metrics --- normalized mutual information (NMI) and adjusted
rand index (ARI) --- in Table~\ref{tab:biencoder_discovery}.
We compare the graph clustering-based inference procedures, which utilize entity information, to one that only uses 
the mention-mention similarities. We
select the hyperparameters, ${k}$ and $\lambda$, on the dev set.

The results indicate that 
the \ours training procedure achieves 
significantly better ARI score compared to the \inbatch
Negatives training on both datasets (10 points on MedMentions 
and 5 points on \zeshel). The 
representations from our proposed training procedure seem to
provide more meaningful clusters of mentions 
than simply taking the layer before the softmax of a trained linking model (i.e., the \inbatch and k-NN settings). 
We hypothesize that the improvements are achieved because the
Arboresence-based training is closely aligned with the clustering inference procedure. We further observe that the \textsc{directed}
approach offers significant improvement over  \textsc{undirected}  on \zeshel.

\subsection{Experiment Details}
Our experiments are run on top of BLINK~\cite{wu2019zero}, a PyTorch~\cite{paszke2019}
implementation of the bi-encoder architecture for entity linking. Each training procedure is run on a single machine using 2 NVIDIA Quadro RTX 8000 GPUs. Our models for Zeshel and MedMentions have 218M and 230M parameters, respectively. Each variant of our bi-encoder models is optimized using mini-batch gradient descent using the Adam optimizer for 5 epochs using a mini-batch size of 128 to accumulate the gradients. Experiments with batch sizes < 128 performed poorly possibly due to increased fluctuation of gradients, and sizes > 128 were computationally infeasible to run given our resources. For \zeshel, each model is trained using 192 warm-up steps and learning rates of 1e-5, 3e-5, and 3e-5 for \inbatch, k-NN, and Arborescence-based models, respectively. For MedMentions, each model is trained using 464 warm-up steps and a learning rate of 3e-5. We use FAISS\footnote{https://github.com/facebookresearch/faiss}~\cite{JDH17} 
for performing k-NN search during graph construction for training and inference. For MedMentions, this took 70 mins to embed and index 2M entities and 120K mentions, and 20 mins to perform search for 120K mentions.

\section{Related Work}

\paragraph{Entity Linking} Entity linking has been
widely studied \cite[inter alia]{milne2008learning,cucerzan-2007-large,lazic2015plato,gupta-etal-2017-entity,raiman2018deeptype,kolitsas-etal-2018-end,decao2020autoregressive}.
Apart from \citet{angell2021clusteringbased}, we note the similarities
of our work and that of \citet{dutta-weikum-2015-c3el}, which 
combines clustering-based cross-document coreference decisions and 
linking. However, \citet{dutta-weikum-2015-c3el} 
is based around using sparse bag-of-word representations and is not well suited for the embedded-based
representations used in this work.  \citet{hoffart-etal-2011-robust,cheng-roth-2013-relational,ganea-hofmann-2017-deep,le-titov-2018-improving} use global objectives instead of independent predictions,  measuring the compatibility
of entity links. Contemporaneous work by \citet{jiang2022towards} also explores a joint paradigm by building clusters of mentions and entities in order to solve a maximum spanning tree problem.

\paragraph{Cross-document Coreference} Models have 
also been developed for the cross-document 
coreference setting where no entity KB is assumed in 
advance \cite[inter alia]{bagga-baldwin-1998-entity-based,gooi-allan-2004-cross,singh-etal-2011-large,barhom-etal-2019-revisiting,cattan2020streamlining,caciularu2021cross,ravenscroft2021cd2cr,cattan2021scico,loganbenchmarking}. In future work, one might explore using our proposed approach for entity discovery in these settings to understand room for improvement of such systems using entity KB information. 

\paragraph{Alternatives to Cross-Encoders} Our work demonstrates how clustering-based training and prediction improves bi-encoder based models for linking and discovery. If prediction efficiency, and not training efficiency, was the only concern, one could use model distillation \cite[inter alia]{hinton2015distilling,izacard2021distilling}. We could also consider models such as poly-encoders as an alternative to bi-encoders \cite{Humeau2020Poly-encoders:}.

\paragraph{Ultrametric Fitting \& Supervised Clustering} The proposed training objective is related to supervised clustering \cite{pmlr-v97-yadav19a}. Algorithms for minimizing the minimax path distance is closely related to fitting ultrametrics \cite[inter alia]{DBLP:conf/nips/ChierchiaP19,pmlr-v119-cohen-addad20a}.

\paragraph{Loss Augmented Inference} Our proposed training objective
uses our graph-clustering procedure to determine the loss. This is 
reminiscent structured prediction approaches such as structured SVMs 
\cite{tsochantaridis2004support}, in which inference is  run during training time and 
the loss is a function of predicted and target structures. 

\section{Conclusion}

In this work, we presented novel arboresence-based training and inference procedures of bi-encoder models for entity linking and discovery. 
Our results indicate that our training procedure yields bi-encoder models which far outperform those models trained with standard procedures on challenging entity linking and entity discovery tasks. Future work includes using cross-encoder models, model distillation, and loss functions for directly optimizing clustering metrics in entity discovery.
\section{Ethical Considerations}

The base models, which we fine-tuned, 
and evaluation datasets are all publicly available.
We will also make our code and models publicly available. 
The task of entity resolution and discovery is relatively
innocuous. However, there are several ways in which 
models could be biased and there is the potential 
for those biases to have harmful downstream consequences. 
There is a large body of work studying the biases of 
language models (such as those used for fine-tuning here)
and coreference models. Most notably in understanding when
error rates in coreference differ across certain populations
(e.g., genders, races, or any entity-type more broadly). 
If entity linking and discovery systems are used to build /
populate knowledge-bases, those systems may propagate these
biased predictions. This could be particularly problematic
if one used such a biased knowledge-base with this realization.
For instance, if entity mentions are author names on citation data
and the entities are scientific authors, statistics like h-index or
citation count could be biased if the algorithms used to disambiguate
the author names are biased. Lastly, we note 
entity linking and discovery are related to 
surveillance and tracking in computer vision, which bear
a substantial weight of ethical considerations.

\section*{Acknowledgements}

We thank Nishant Yadav, Sunil Mohan, and members of UMass IESL and NLP groups for helpful discussion and feedback.
This work is funded in part by the Center for Data Science and the Center for Intelligent Information
Retrieval, and in part by the National Science Foundation under Grants No. 1763618, and in part by the Chan Zuckerberg Initiative under the project Scientific Knowledge
Base Construction. 
The work reported here was supported in part by the Center for Data Science and the Center for Intelligent Information Retrieval, and in part using high performance computing equipment obtained under a grant from the Collaborative R\&D Fund managed by the Massachusetts Technology Collaborative.
Rico Angell is supported by the National Science Foundation Graduate Research Fellowship under Grant No. 1938059.
Any opinions, findings and conclusions or recommendations expressed in this
material are those of the authors and do not necessarily reflect those of the sponsor.

\bibliography{anthology,custom}
\bibliographystyle{acl_natbib}

\end{document}